\documentclass[journal]{IEEEtran}
\usepackage{cite}
\ifCLASSINFOpdf
\else
\fi
\usepackage{mathrsfs}
\usepackage{amsfonts}
\usepackage{mathdots}

\usepackage{booktabs}
\usepackage{threeparttable}
\usepackage{multirow}
\usepackage{array}
\usepackage{graphicx}
\usepackage{epstopdf}
\usepackage{float}
\usepackage{stfloats}
\usepackage{subfigure}

\usepackage{amsmath}
\usepackage{algorithm}
\usepackage{algorithmic}
\usepackage{array}

\usepackage{stfloats}

	\usepackage{booktabs}
	\usepackage{multirow}
	\usepackage{array}
	\usepackage{graphicx}
	\usepackage{epstopdf}
	\usepackage{float}
	\usepackage{stfloats}
	\usepackage{mathrsfs}
	\usepackage{amsfonts}
	\usepackage{mathdots}
	
	\usepackage{graphicx}
	\usepackage{footmisc}

\begin{document}
	
\title{LINGO: Latent Initialization and Gradient Optimization for Sparse-view X-ray Novel View Synthesis and CT Reconstruction with 3D Gaussian Splatting}

\author{Lifeng Xing, Dequan Jin, Kunpeng Bu, Peigeng He and~Shihui~Ying~\IEEEmembership{Member,~IEEE}
 
\thanks{This work was supported in part by the National Natural Science Foundation of China under Grant 12531019, and the special foundation for Guangxi Ba Gui Scholars. (Corresponding author: Dequan Jin).}
\thanks{L. Xing and D. Jin are with School of Mathematics and Center for Applied Mathematics of Guangxi, Guangxi University, Nanning 530004, China (email: dqjin@gxu.edu.cn, ayacyann@st.gxu.edu.cn).}
\thanks{K. Bu and P. He are with Department of Endocrinology, Metabolism and Nephrology, Guangxi Medical University Cancer Hospital, Nanning 530021, China (email: bukunpeng2008@163.com, colxh7@163.com).}
\thanks{S. Ying is with School of Mechanics and Engineering Science and Shanghai Institute of Applied Mathematics and Mechanics, Shanghai University, Shanghai 200072, China(e-mail: shying@shu.edu.cn).}
}
\markboth{Journal of \LaTeX\ Class Files,~Vol.~14, No.~8, August~2015}%
{Shell \MakeLowercase{\textit{et al.}}: Bare Demo of IEEEtran.cls for IEEE Journals}

\maketitle

\begin{abstract}

In novel view synthesis and Computed Tomography (CT) reconstruction with sparse-view X-ray imaging, insufficient angular coverage leads to structural ambiguity and accumulated noise. Integrating 3D Gaussian Splatting (3DGS) with X-ray absorption physics can achieve promising results, but it suffers from noisy initialization, positional insensitivity, and weak gradients in low-density regions. In this paper, we propose a unified Latent Initialization and Gradient Optimization (LINGO) framework to address these issues. LINGO combines latent mask-space initialization with dynamic gradient optimization to improve point cloud structural completeness while accelerating training. It constructs voxel-level 3D filters from X-ray masks to robustly suppress background noise and provide reliable geometric priors. By employing an adaptive voxel scaling strategy and dynamically scaling loss, LINGO can adjust spatial resolution and explicitly amplify gradients in low-density structures. To evaluate the quality of initialization, we introduce the Initialization Point Cloud Structural Deviation (IPSD) metric. Experiments on the X3D dataset indicate that for the novel view synthesis task, LINGO improves the Peak Signal-to-Noise Ratio (PSNR) and Structural Similarity Index (SSIM) by an average of 0.72 and 0.0039, respectively, over baselines under identical sparse-view settings, achieving comparable reconstruction quality within 5k steps to state-of-the-art models typically trained with 30k iterations. For the CT reconstruction task, LINGO also demonstrates consistent improvements, with average PSNR/SSIM gains of 0.36/0.0134. These results highlight LINGO’s effectiveness in both accelerating training and enhancing reconstruction quality across different sparse-view imaging scenarios.

\end{abstract}


\section{Introduction}\label{sec:introduce}

In medical imaging, Computed Tomography (CT) reconstruction\cite{ct-principle,ct-tech,ct-med} is a pivotal technique for disease diagnosis and clinical assessment. However, conventional multi-view acquisition protocols require a large number of projections, which inevitably expose patients to substantial radiation doses. To address this issue, sparse-view CT aims to reduce radiation exposure by limiting the number of projection views. Despite its clinical advantages, this setting introduces severe reconstruction artifacts and structural degradation due to insufficient angular sampling and incomplete information.

To address these issues, 3D Gaussian Splatting (3DGS)\cite{3dgs} has emerged as an efficient explicit 3D representation\cite{EWA} for sparse-view CT reconstruction. 3DGS represents scenes as a collection of Gaussian primitives with both spatial and radiative attributes.  Since it is naturally suitable for sparse-view reconstruction tasks due to its inherent sparsity and continuous representation capability, $\text{R}^2$-Gaussian\cite{r2_gaussian} incorporates X-ray absorption physics\cite{ct-principle} into its framework and enables physically grounded modeling of X-ray formation, demonstrating excellent performance in CT reconstruction scenarios\cite{ct-reconstruction}.

Despite the promising performance achieved by  $\text{R}^2$-Gaussian\cite{r2_gaussian}, several challenges remain in its application to sparse-view CT reconstruction. First, the lack of rich textures and distinct structural cues in X-ray images inherently complicates the extraction of reliable feature correspondences\cite{colmap}. An available method for this issue relies on a simple initialization pipeline consisting of Feldkamp–Davis–Kress (FDK)\cite{FDK} back-projection followed by threshold-based denoising. Nevertheless, this scheme cannot effectively suppress widespread background noise, leading to an initial Gaussian point cloud contaminated by severe interference and outliers. These artifacts further degrade the quality of subsequent optimization and compromise its convergence stability. Second, $\text{R}^2$-Gaussian exhibits limited sensitivity to spatial position, leading to insufficient positional updates and weak structural refinement during optimization. This issue arises from the nature of the X-ray imaging model\cite{ct-tech}, where all points along a projection ray contribute cumulatively to the measured intensity. As a result, positional perturbations along the ray direction produce only marginal changes in the projection for a single view, weakening the optimization signal for spatial adjustment. Moreover, projections from different viewpoints may induce conflicting gradient directions for the same Gaussian primitive (e.g., structural versus non-structural regions), which can lead to gradient cancellation and make it difficult to suppress noise points or sharpen structural boundaries. Finally, low-density structures (e.g., small vessels or aneurysms) yield weak loss signals, resulting in insufficient gradients, slow convergence, and poor recovery of fine details.

To address these challenges, we propose Latent Initialization and Gradient Optimization (LINGO). We address the challenges aforementioned from the following two aspects: (1) To suppress the background noises, we construct voxel-level 3D filters from X-ray masks; (2) To increase the sensitivity of spatial position, we provide reliable geometric priors and then employ an adaptive voxel scaling strategy and dynamically scaling loss to adjust spatial resolution and explicitly amplify gradients in low-density structures. The main contributions of this paper are as follows:

\begin{itemize}
   \item This paper proposes a latent mask-space initialization strategy that constructs voxel-level filters from X-ray masks, yielding a cleaner and more structurally consistent initial Gaussian point cloud.
  \item This paper introduces a dynamic gradient optimization scheme with a loss-scaling strategy and voxel scaling mechanism, to enhance gradient responses in low-density regions and reduce computational overhead.
  \item This paper provides the IPSD metric to evaluate the initialization quantitatively.  
  \item Comparative experiments and ablation study demonstrate the advantages of LINGO in novel view synthesis and CT reconstruction under sparse-view conditions.
\end{itemize}

\section{Related work}\label{sec:relatedwork}

In the domain of CT reconstruction\cite{ct-reconstruction}, the FDK\cite{FDK} algorithm is the most widely adopted filtered back-projection method for cone-beam geometry\cite{cone-beam-ct}, renowned for its computational efficiency, which enables the rapid reconstruction of 3D volumetric images from multi-angle projection data. However, FDK degrades severely under sparse-view settings. Methods including Simultaneous Algebraic Reconstruction Technique (SART)\cite{SART}, Adaptive Sparsity Domain Reconstruction (ASD-POCS)\cite{ASD-POCS}, and Conjugate Gradient Least Squares (CGLS)\cite{cgls} are proposed to overcome the inherent limitations of analytical methods, a series of algebraic iterative reconstruction techniques\cite{os}. These methods iteratively alternate between projection and volume spaces, integrating regularization terms and prior knowledge to enhance reconstruction quality under undersampled conditions. Nevertheless, these iterative approaches remain limited by insufficient sample and high computational cost.

With the rapid development of neural representations\cite{NeRF,3dgs}, medical image reconstruction has evolved from direct low-dose CT reconstruction\cite{low-doseCT,low-dose-GAN,TomoGAN,SparseView2DCT,Med-NeRF} to a two-stage pipeline\cite{NeRF,NAF,sax_nerf,3dgs,x_gaussian,r2_gaussian}, where sparse X-ray projections are first used to synthesize missing views, followed by back-projection to recover volumetric density. This paradigm leverages neural rendering to compensate for incomplete data, reducing the dependence of traditional reconstruction methods\cite{FDK,SART,ASD-POCS,cgls} on dense sampling and enabling more accurate reconstruction under sparse-view conditions. Among these approaches, Neural Radiance Field (NeRF)-based methods\cite{NeRF,Zip-NeRF,Mip-NeRF360,Mip-NeRF,Med-NeRF,sax_nerf} represent scenes as continuous volumetric radiance fields with high-quality novel view synthesis via differentiable rendering. Extensions such as SAX-NeRF\cite{sax_nerf} further adapt this framework to X-ray imaging. However, despite various acceleration techniques\cite{Eff-NeRF,NerfAcc,TensoRF,MERF,DDRs}, NeRF still suffers from high computational cost and slow updates due to its dense volumetric representation, limiting its applicability to high-resolution and real-time CT reconstruction.

3DGS\cite{3dgs} offers a more efficient explicit representation by representing scenes as a set of Gaussian primitives\cite{EWA} with position, covariance, and opacity. It enables rapid, differentiable rendering and optimization, achieving significant breakthroughs not only in general computer vision fields such as 3D scene generation\cite{3dgs,LucidDreamer,HUGSHG,humanarticulatedgaussiansplatting,gs2mesh,Scaffold-GS,GaussianShader}, dynamic scene modeling\cite{4dgs,Dynamic3G,PhysGaussian,4dgs-Photorealistic}, simultaneous localization and mapping\cite{slam,GS-SLAM}, and inverse rendering\cite{GSIR}, but also in X-ray reconstruction\cite{x_gaussian,r2_gaussian}. The X-Gaussian\cite{x_gaussian} is the first method to apply 3DGS to X-ray image reconstruction. $\text{R}^2$-Gaussian\cite{r2_gaussian} further combines 3DGS with X-ray absorption laws to achieve more accurate X-ray novel view synthesis and CT reconstruction. 

Compared to implicit neural rendering methods like NeRF, 3DGS provides a sparser and geometrically more explicit scene representation with significantly faster convergence speeds and a more intuitive geometric interpretation. These advantages provide 3DGS with immense potential in X-ray CT reconstruction tasks, enabling not only efficient view synthesis but also providing a physically meaningful point cloud structure representation, thus offering a powerful tool for high-precision medical image reconstruction under sparse-view conditions. However, the initialization strategy of $\text{R}^2$-Gaussian is too simple, combining FDK\cite{FDK} back-projection with threshold denoising. This strategy cannot eliminate pervasive residual noise in the background, leading to interference signals in the initial Gaussian point clouds. $\text{R}^2$-Gaussian is also insensitive to position, so that it cannot effectively remove Gaussian points at structural boundaries or noise regions, or guide them toward accurate structural reconstruction. Moreover, the loss for low-density structural regions is too small, resulting in weak gradients and slow convergence. 

\section{Preliminaries}\label{sec:prelim}

\subsection{Otsu’s Method}

Otsu’s method is a statistical image segmentation approach\cite{Otsu}. Given an observed X-ray projection image $\mathcal{I}_o \in \mathbb{R}^{H \times W}$, where each pixel represents the measured intensity, the method searches over all possible grayscale thresholds $t$ and selects the optimal threshold 
\begin{equation}
	t^{\star} =\text{Otsu}(\mathcal{I}_o),
\end{equation}
to maximizes the intra-class variance. This enables automatic separation of foreground and background with high stability and computational efficiency.

% Otsu’s method is a statistical image segmentation method\cite{Otsu}. Its core idea is to search through all possible grayscale thresholds and select the one that maximizes the between-class variance, thereby achieving automatic separation of foreground and background with computational simplicity and high stability. 

% Suppose the grayscale range of an image $\mathcal{I}_o$ is $[0, L-1]$. Let $p_i$ be the probability of pixels with grayscale level $i$. Denote the background class and foreground class of $\mathcal{I}_o$ for a given threshold $t$ by $$C_0: 0 \le i \le t,$$ and $$C_1: t + 1 \le i \le L - 1.$$ Calculate the sum of probabilities for $C_0$ and $C_1$:
% \begin{equation}
%   w_0 = \sum_{i=0}^{t} p_i, \quad w_1 = \sum_{i=t+1}^{L-1} p_i,
% \end{equation}
% and their mean intensities are given by:
% \begin{equation}
%   \mu_0 = \frac{\sum_{i=0}^{t} i p_i}{w_0}, \quad \mu_1 = \frac{\sum_{i=t+1}^{L-1} i p_i}{w_1}.
% \end{equation}
% The total mean intensity of the image is given by:
% \begin{equation}
%   \mu_T = \sum_{i=0}^{L-1} i p_i,
% \end{equation}
% and then the between-class variance is formulated by:
% \begin{equation}
%   \sigma_b^2 = w_0 (\mu_0 - \mu_T)^2 + w_1 (\mu_1 - \mu_T)^2.
% \end{equation}
% Since maximizing the between-class variance results in the largest distinction between the two classes, leading to the optimal segmentation, then the optimal threshold $t^{\star}$ is given by:
% \begin{equation}
%   t^{\star} = \arg \max_{t} \sigma_b^2(t).
% \end{equation}

%We denote the whole process as:

\subsection{3D Gaussian Splatting}

3DGS represents a scene as a set of Gaussian primitives and performs rendering based on the volume rendering formulation\cite{EWA}. The color along a ray is obtained via density accumulation and transmittance, and can be discretized as:
\begin{equation}
C = \sum_{i=1}^{N} T_i \alpha_i c_i,\quad T_i = \prod_{j=1}^{i-1}(1 - \alpha_j).
\end{equation}
The opacity $\alpha_i$ is parameterized by a 3D Gaussian function:
\begin{equation}
	\begin{aligned}
\alpha_i &=  G_i^3(\mathbf{x} | \hat{\alpha}_i, \mathbf{p}_i, \mathbf{\Sigma}_i)\\&= \hat{\alpha}_i \exp\left(-\frac{1}{2}(\mathbf{x} - \mathbf{p}_i)^{T}\mathbf{\Sigma}_i^{-1}(\mathbf{x} - \mathbf{p}_i)\right),
\end{aligned}
\end{equation}
where $\mathbf{p}_i$ and $\mathbf{\Sigma}_i$ are the central position and covariance.

3DGS relies on transparency-based compositing\cite{alpha-blending} that differs from the X-ray imaging process governed by attenuation physics. To bridge this gap, $\text{R}^2$-Gaussian reformulates the rendering process based on the Beer–Lambert law\cite{r2_gaussian}:
\begin{equation}
I = \int_{t_n}^{t_f}\sigma(\mathbf{r}(t))dt.
\end{equation}
The volumetric density is the sum of Gaussian functions:
\begin{equation}
\sigma(\mathbf{x}) = \sum_{i=1}^{M} G_i^3(\mathbf{x} | \rho_i, \mathbf{p}_i, \mathbf{\Sigma}_i),
\end{equation}
where $\rho_i$ is the central density. 

A key property of Gaussian distributions is that integrating a normalized 3D Gaussian function along a ray yields a 2D Gaussian function. Thus, the rendering equation can be simplified as:
\begin{equation}
  \begin{aligned}
    I & = \int\sum_{i=1}^{M} G_i^3(\mathbf{x} | \rho_i, \mathbf{p}_i, \mathbf{\Sigma}_i)dt \\
    & = \sum_{i=1}^{M} G_i^2(\mathbf{\hat{x}} | \sqrt{\frac{2\pi |\mathbf{\Sigma}_i|}{|\mathbf{\hat{\Sigma}}_i|}}\rho_i, \mathbf{\hat{p}}_i, \mathbf{\hat{\Sigma}}_i).
  \end{aligned}
\end{equation}
Here, $\mathbf{\hat{x}},\, \mathbf{\hat{p}}_i,\, \mathbf{\hat{\Sigma}}_i$ denote the corresponding quantities projected from 3D space onto the 2D image space.

\begin{figure*}[hbt]
\centering
    \includegraphics[width=\textwidth]{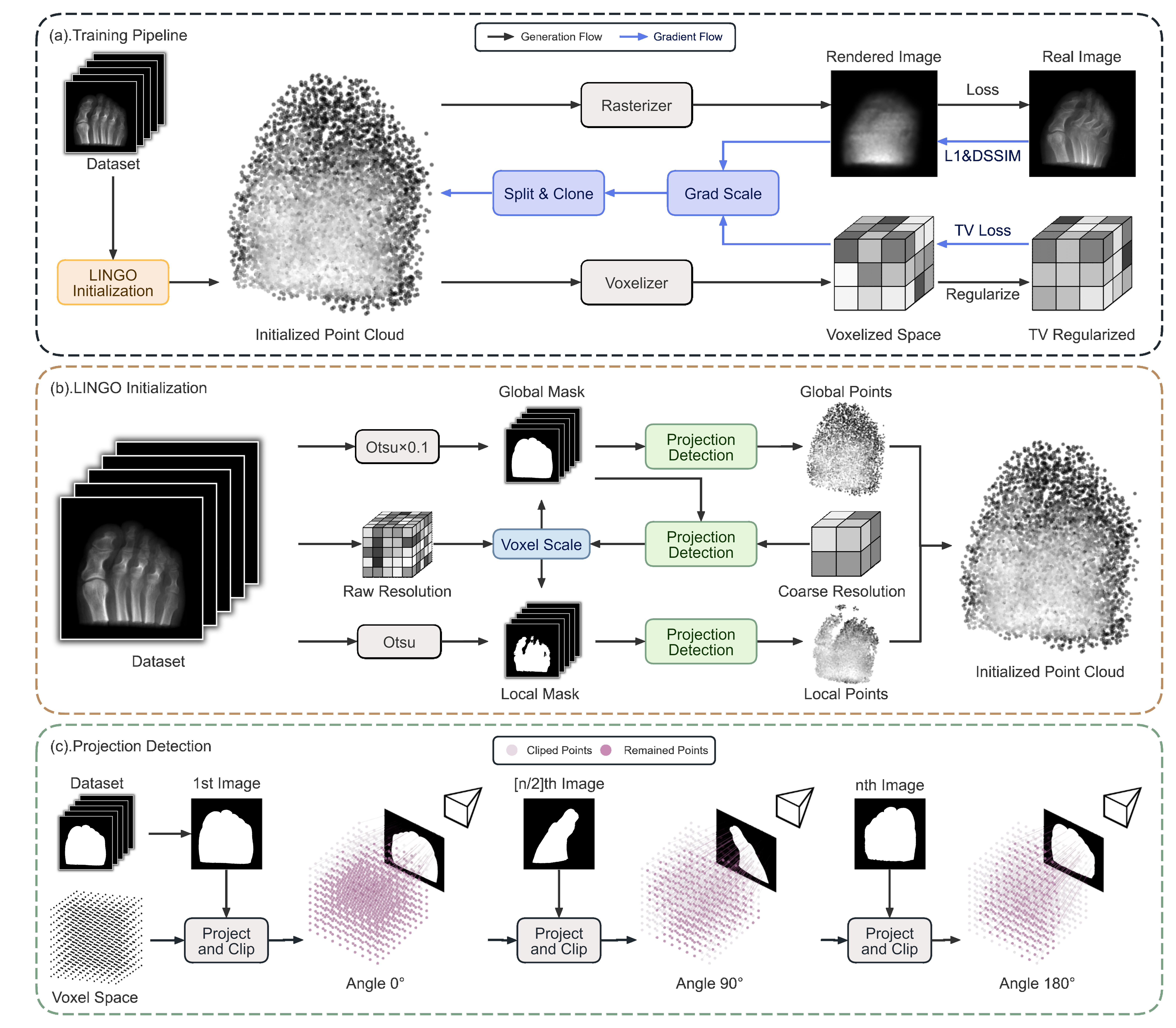}
    \caption{Overview of the proposed LINGO framework. (a) Overall training pipeline of LINGO with $\text{R}^2$-Gaussian backbone. (b) LINGO initialization module for mask-guided Gaussian point generation. (c) Mask detection and filtering submodule within the initialization stage.}
    \label{fig:flowchart} 
\end{figure*}

\section{Method}\label{sec:method}

In this section, we present the proposed LINGO framework.  We first develop a robust initialization strategy to suppress background noise and improve the quality of the initial Gaussian point set. Then we introduce a dynamic loss-scaling strategy to enhance gradient responses in low-density regions during optimization. A new metric, Initialization Point Cloud Structural Deviation (IPSD), is proposed to quantitatively evaluate how well the initial point set approximates the underlying structural distribution. The overall pipeline is illustrated in Fig.~\ref{fig:flowchart}.

% In this section, we present the proposed LINGO framework, as illustrated in Fig.~\ref{fig:flowchart}. We first develop a robust initialization strategy by incorporating X-ray mask information into a voxel-level filtering process during back-projection, effectively suppressing background noise and improving the quality of the initial Gaussian point set. Building upon this, we introduce a dynamic loss-scaling strategy to enhance gradient responses in low-density regions during optimization. Finally, we propose the Initialization Point Cloud Structure Deviation (IPSD) error to quantitatively evaluate how well the initial point set approximates the underlying structural distribution.

\subsection{Mask-guided Voxel Filtering}\label{sub:mask_projection_detection}

To construct a voxel-level filtering process, we define a voxel resolution $\mathcal{V}_d = (V_1, V_2, V_3)$ which contains $N = V_1 \times V_2 \times V_3$ points in total. For each X-ray image $\mathcal{I}_o$ where $o = 1, 2, \dots, n$, we first determine a high-density threshold 
\begin{equation}
  T_{o}^h = \text{Otsu}(\mathcal{I}_o),
\end{equation}
by Otsu's method\cite{Otsu} to distinguish high-density regions from low-density regions in the image, and then a low-density threshold
\begin{equation}
  T_{o}^l = T_{o}^h \times 0.1,
\end{equation}
to ensure that low-density areas are also considered during the initialization phase. The two thresholds have different functions in initialization. \(T_{o}^l\) primarily removes zero-value background and low-intensity noise, while \(T_{o}^h\) focuses on extracting prominent high-density regions.

We use these thresholds to generate two binary mask images for each X-ray image $\mathcal{I}_o$:
\begin{equation}
\mathcal{M}^{{\mathcal{I}}_l}_o = (\mathcal{I}_o > T_{o}^l), 
\quad
\mathcal{M}^{{\mathcal{I}}_h}_o = (\mathcal{I}_o > T_{o}^h).
\end{equation}
They represent the global structure, including low-density regions and the local structure of high-density regions, respectively.

Denote the points in the voxel-space by $\mathbf{P}_i^{\mathcal{V}}$ ($i = 0, 1, \dots, N$). We project each $\mathbf{P}_i^{\mathcal{V}}$ onto every image plane using the camera projection matrix $\mathbf{M}_o$ for an image $\mathcal{I}_o$ and obtain its image-space coordinates:
\begin{equation}
\mathbf{P}^{\mathcal{I}_o} = \mathrm{floor}\left(\mathrm{clip}(\mathbf{M}_o \mathbf{P}^{\mathcal{V}})\right).
\end{equation}
Here $\mathrm{clip}(\cdot)$ removes points outside the image bounds, and the operation $\mathrm{floor}(\cdot)$ is applied to prevent valid points from being erroneously excluded due to upward rounding.

To retain the points projecting within the masked regions across all views, for each projected point $\mathbf{P}^{\mathcal{I}_o}$, we check if it falls within $\mathcal{M}^{{\mathcal{I}}_l}_o$ and $\mathcal{M}^{{\mathcal{I}}_h}_o$ by the low-density and high-density masks:
\begin{equation}
\mathcal{M}^{\mathbf{P}_l}_o = \mathcal{M}^{\mathcal{I}_l}_o[\mathbf{P}^{\mathcal{I}_o}], 
\quad
\mathcal{M}^{\mathbf{P}_h}_o = \mathcal{M}^{\mathcal{I}_h}_o[\mathbf{P}^{\mathcal{I}_o}].
\end{equation}
We repeat this process for all images and obtain the valid point masks $\mathcal{M}^\mathbf{P}_l$ and $\mathcal{M}^\mathbf{P}_h$ by:
\begin{equation}
\mathcal{M}^\mathbf{P}_l = \bigcap_{o=1}^{n}\mathcal{M}^{\mathbf{P}_l}_o, 
\quad
\mathcal{M}^\mathbf{P}_h = \bigcap_{o=1}^{n}\mathcal{M}^{\mathbf{P}_h}_o.
\end{equation}
These masks are used to obtain the valid voxel-space point sets:
\begin{equation}\label{eq:pv}
\mathbf{P}^{\mathcal{V}_l}_{\text{valid}} = \mathbf{P}^\mathcal{V}[\mathcal{M}^{\mathbf{P}}_l], 
\quad
\mathbf{P}^{\mathcal{V}_h}_{\text{valid}} = \mathbf{P}^\mathcal{V}[\mathcal{M}^{\mathbf{P}}_h].
\end{equation}
To increase voxel resolution within high-density regions $\mathbf{P}^{\mathcal{V}_h}_{\text{valid}}$, we apply a local interpolation 
\begin{equation}
\hat{\mathbf{P}}^{\mathcal{V}_h}_{\text{valid}} = {\mathbf{P}}^{\mathcal{V}_h}_{\text{valid}} + (\frac{i}{V_1},\frac{j}{V_2},\frac{k}{V_3}),
\end{equation}
for $i,j,k\in [-0.5,0.5]$ to generate a finer point cloud in these areas.  

The entire mask projection and filtering process effectively removes background noise and provides precise initialization priors for subsequent Gaussian parameter initialization.

\subsection{Voxel Scaling for Spatial Occupancy}\label{sub:voxel_scale}

Before initializing Gaussian parameters, we precompute a spatial occupancy ratio to scale the voxel space adaptively. This step is essential for computational efficiency and ensures an effective distribution of Gaussian points.

We begin with a coarse voxel resolution \(\mathcal{V} = (V,V,V)\) to estimate the spatial occupancy of the target structure. Based on the valid low-density point set \(\mathbf{P}^{\mathcal{V}_l}_{\text{valid}}\) obtained from Eq.~(\ref{eq:pv}), we first compute the occupancy ratio \(P\) as:
\begin{equation}
P = \frac{\text{count}(\mathbf{P}^{\mathcal{V}_l}_{\text{valid}})}{V^3},
\end{equation}
where $\text{count}(\cdot)$ denotes the number of valid points.

Using the estimated occupancy ratio, we define the scaling factor \(S\) as:
\begin{equation}
	S = \max(P V_S, 1),
\end{equation}
where $V_S$ is a predefined upper bound controlling the maximum voxel scaling. This formulation ensures that the scaling factor remains stable while preventing excessive downsampling. With a given the original device voxel resolution $\mathcal{V}_d = (V_1, V_2, V_3)$, the scaled voxel resolution is presented by:
\begin{equation}
	\mathcal{V}_{d_s} = \left(\frac{V_1}{S}, \frac{V_2}{S}, \frac{V_3}{S}\right).
\end{equation}

This adaptive voxel scaling strategy reduces computational cost by lowering the effective resolution in dense regions, while simultaneously enlarging voxel sizes to group both structural and noisy points within the same spatial cell. Such grouping facilitates subsequent Gaussian splitting and cloning operations, enabling more effective suppression of non-structural points and improving optimization stability.

% This adaptive voxel scaling strategy reduces computational overhead by lowering the effective resolution in dense regions, while enlarging voxel sizes to group both structural and noisy points within the same spatial cell. Such grouping facilitates subsequent Gaussian splitting and cloning operations, enabling more effective suppression of non-structural points and improving optimization stability.

% \subsection{Initialization of Point Cloud Attributes}

With Gaussian positions determined from mask projection and voxel scaling, we initialize their attributes, particularly density, based on the physical rules of X-ray imaging. According to the Beer-Lambert law, we have
\begin{equation}
	I_o = I_i e^{-\int_{L}\sigma_t\,dt},
\end{equation}
where $I_o$ and $I_i$ denote the output and incident intensities, and $\sigma_t$ is the attenuation coefficient along ray $L$. Taking the logarithm yields:
\begin{equation}
	p = -\ln\left(\frac{I_i}{I_o}\right) = \int_{L}\sigma_t\,dt.
\end{equation}

For back-projection initialization, we assume a constant attenuation $\sigma_t=\sigma$ and a unit ray length ($L=1$), leading to:
\begin{equation}
p = \sigma.
\end{equation}
Thus, each pixel intensity directly corresponds to the attenuation coefficient along the ray. The initial density of voxel-space points, $C^{\mathcal{V}_{d_s}}$, is computed by averaging the pixel intensities at the projected locations across all views:
\begin{equation}
C^{\mathcal{V}_{d_s}} = \frac{1}{n} \sum_{o=1}^{n} \mathcal{I}_o[\mathbf{P}^{\mathcal{I}_o}].
\end{equation}

Compared to filtered back-projection (e.g., FDK), this direct assignment better preserves the original density distribution and avoids smoothing artifacts,  inducing more accurate structural reconstruction.

\begin{table*}[htbp]
\centering
\caption{Quantitative results of X-ray novel view synthesis on the X3D dataset under different numbers of training views.}
\begin{tabular}{|c|c|ccc|ccc|ccc|ccc|}
\hline
 \multirow{2}{*}{View Num} & \multirow{2}{*}{Iter} & \multicolumn{3}{c|}{{3DGS}} & \multicolumn{3}{c|}{{X-Gaussian}} & \multicolumn{3}{c|}{{$\text{R}^2$-Gaussian}} & \multicolumn{3}{c|}{{LINGO}} \\[0.3ex]
\cline{3-14}
  & & {PSNR$\uparrow$} & {SSIM$\uparrow$} & {IPSD$\downarrow$} & {PSNR$\uparrow$} & {SSIM$\uparrow$} & {IPSD$\downarrow$} & {PSNR$\uparrow$} & {SSIM$\uparrow$} & {IPSD$\downarrow$} & {PSNR$\uparrow$} & {SSIM$\uparrow$} & {IPSD$\downarrow$} \\[0.3ex]
\hline
\multirow{3}{*}{5} & 5k & 28.74 & 0.8871 & 0.1241 & 28.88 & 0.8899 & 0.1236 & \underline{32.90} & \underline{0.9449} & \underline{0.1228} & \textbf{33.33} & \textbf{0.9506} & \textbf{0.0282} \\[0.3ex]
                   & 15k & 28.55 & 0.8887 & 0.1202 & 29.05 & 0.8933 & 0.1192 & \underline{33.23} & \underline{0.9466} & \underline{0.1184} & \textbf{33.60} & \textbf{0.9520} & \textbf{0.0264} \\[0.3ex]
                   & 30k & 28.24 & 0.8820 & 0.1199 & 28.73 & 0.8877 & 0.1190 & \underline{33.41} & \underline{0.9466} & \underline{0.1182} & \textbf{33.67} & \textbf{0.9524} & \textbf{0.0264} \\[0.3ex]
\hline
\multirow{3}{*}{10} & 5k & 33.64 & 0.9387 & 0.1251 & 33.64 & 0.9391 & 0.1244 & \underline{37.87} & \underline{0.9699} & \underline{0.1182} & \textbf{38.85} & \textbf{0.9747} & \textbf{0.0275} \\[0.3ex]
                    & 15k & 34.12 & 0.9425 & 0.1211 & 34.17 & 0.9431 & \underline{0.1187} & \underline{38.40} & \underline{0.9710} & 0.1190 & \textbf{39.37} & \textbf{0.9755} & \textbf{0.0259} \\[0.3ex]
                    & 30k & 34.28 & 0.9418 & 0.1208 & 34.35 & 0.9422 & \underline{0.1184} & \underline{38.61} & \underline{0.9710} & 0.1188 & \textbf{39.55} & \textbf{0.9756} & \textbf{0.0258} \\[0.3ex]
\hline
\multirow{3}{*}{25} & 5k & 37.35 & 0.9740 & 0.1273 & 39.38 & 0.9740 & 0.1303 & \underline{43.82} & \underline{0.9883} & \underline{0.1152} & \textbf{44.55} & \textbf{0.9895} & \textbf{0.0242} \\[0.3ex]
                    & 15k & 40.95 & 0.9781 & 0.1205 & 40.99 & 0.9781 & 0.1246 & \underline{45.49} & \underline{0.9897} & \underline{0.1130} & \textbf{46.25} & \textbf{0.9909} & \textbf{0.0233} \\[0.3ex]
                    & 30k & 41.36 & 0.9777 & 0.1203 & 41.26 & 0.9777 & 0.1242 & \underline{46.09} & \underline{0.9899} & \underline{0.1129} & \textbf{47.04} & \textbf{0.9911} & \textbf{0.0233} \\[0.3ex]
\hline
\end{tabular}
\label{tab:novel_view_synthesis}
\end{table*}

\begin{figure*}[htb]
\centering
    \includegraphics[width=\textwidth]{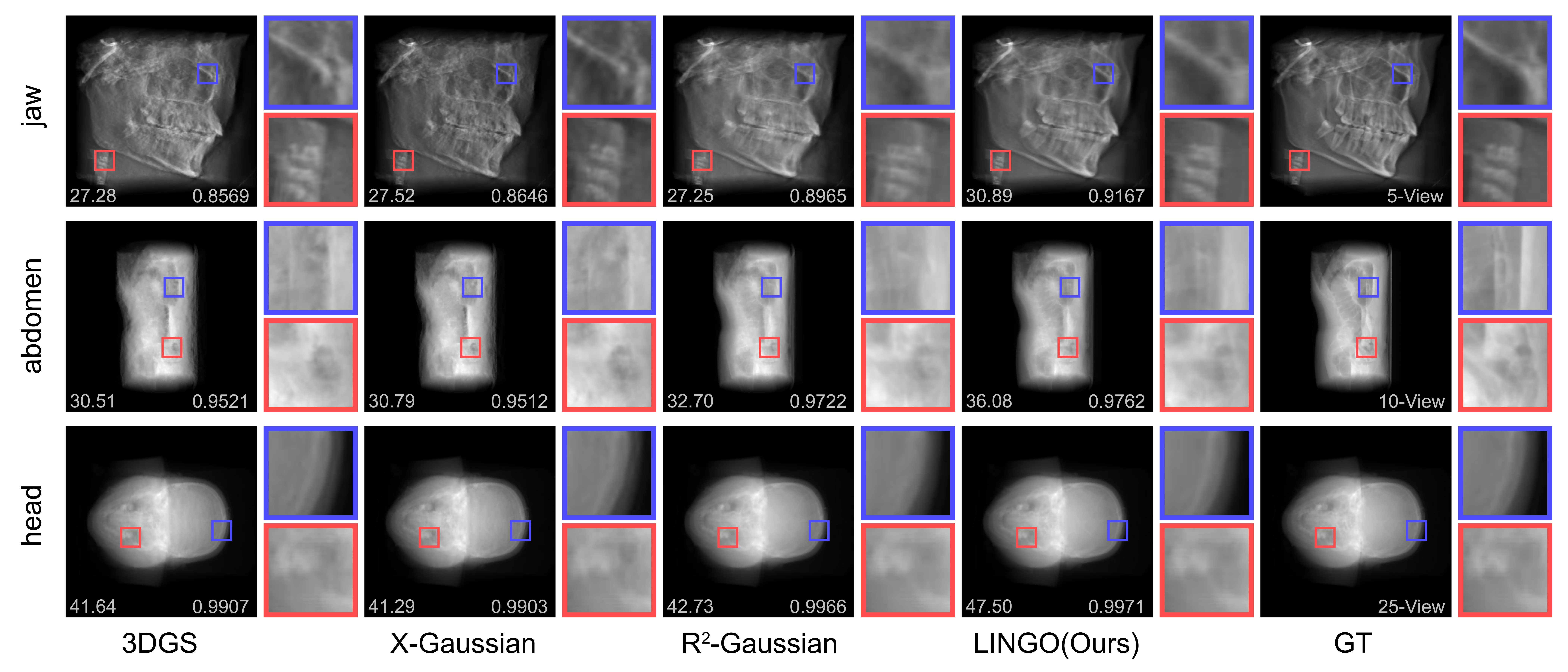}
    \caption{Visualization of results for the novel view synthesis task on X3D dataset.}
    \label{fig:visual_result_xray}
\end{figure*}

\begin{table*}[htbp]
\centering
\caption{Quantitative results of CT reconstruction on the X3D dataset under different numbers of training views.}
\begin{tabular}{|c|c|cc|cc|cc|cc|cc|}
\hline
\multirow{2}{*}{{View number}} & \multirow{2}{*}{{Method}} & \multicolumn{2}{c|}{{None}} & \multicolumn{2}{c|}{{+3DGS}} & \multicolumn{2}{c|}{{+X-Gaussian}} & \multicolumn{2}{c|}{{+$\text{R}^2$-Gaussian}} & \multicolumn{2}{c|}{{+LINGO}} \\[0.3ex]
\cline{3-12}
 & & {PSNR$\uparrow$} & {SSIM$\uparrow$} & {PSNR$\uparrow$} & {SSIM$\uparrow$} & {PSNR$\uparrow$} & {SSIM$\uparrow$} & {PSNR$\uparrow$} & {SSIM$\uparrow$} & {PSNR$\uparrow$} & {SSIM$\uparrow$} \\[0.3ex]
\hline
\multirow{3}{*}{5} & FDK\cite{FDK} & 12.63 & 0.1183 & 20.05 & 0.2600 & 20.11 & 0.2621 & \underline{22.98} & \underline{0.3986} & \textbf{23.29} & \textbf{0.4139} \\[0.3ex]
 & SART\cite{SART} & 23.89 & 0.6614 & 21.38 & 0.5602 & 21.35 & 0.5633 & \underline{24.61} & \underline{0.7025} & \textbf{24.94} & \textbf{0.7202} \\[0.3ex]
 & ASD-POCS\cite{ASD-POCS} & 23.38 & 0.6486 & 22.05 & 0.6113 & 22.18 & 0.6194 & \underline{25.46} & \underline{0.7488} & \textbf{25.78} & \textbf{0.7643} \\[0.3ex]
\hline
\multirow{3}{*}{10} & FDK\cite{FDK} & 16.13 & 0.1745 & 21.94 & 0.3075 & 21.98 & 0.3101 & \underline{24.12} & \underline{0.4153} & \textbf{24.37} & \textbf{0.4246} \\[0.3ex]
 & SART\cite{SART} & 26.22 & 0.7163 & 24.53 & 0.6480 & 24.55 & 0.6536 & \underline{27.28} & \underline{0.7614} & \textbf{27.76} & \textbf{0.7798} \\[0.3ex]
 & ASD-POCS\cite{ASD-POCS} & 25.64 & 0.7215 & 25.98 & 0.7189 & 26.04 & 0.7214 & \underline{28.45} & \underline{0.8066} & \textbf{28.90} & \textbf{0.8252} \\[0.3ex]
\hline
\multirow{3}{*}{25} & FDK\cite{FDK} & 20.81 & 0.2973 & 23.92 & 0.3828 & 23.91 & 0.3825 & \underline{25.18} & \underline{0.4489} & \textbf{25.21} & \textbf{0.4503} \\[0.3ex]
 & SART\cite{SART} & 30.22 & 0.8015 & 28.90 & 0.7717 & 28.91 & 0.7715 & \underline{31.45} & \underline{0.8467} & \textbf{31.75} & \textbf{0.8526} \\[0.3ex]
 & ASD-POCS\cite{ASD-POCS} & 29.43 & 0.8233 & 30.98 & 0.8425 & 31.00 & 0.8422 & \underline{32.61} & \underline{0.8864} & \textbf{32.91} & \textbf{0.8924} \\[0.3ex]
\hline
\end{tabular}
\label{tab:ct_reconstruction}
\end{table*}

\begin{figure*}[htb]
\centering
    \includegraphics[width=\textwidth]{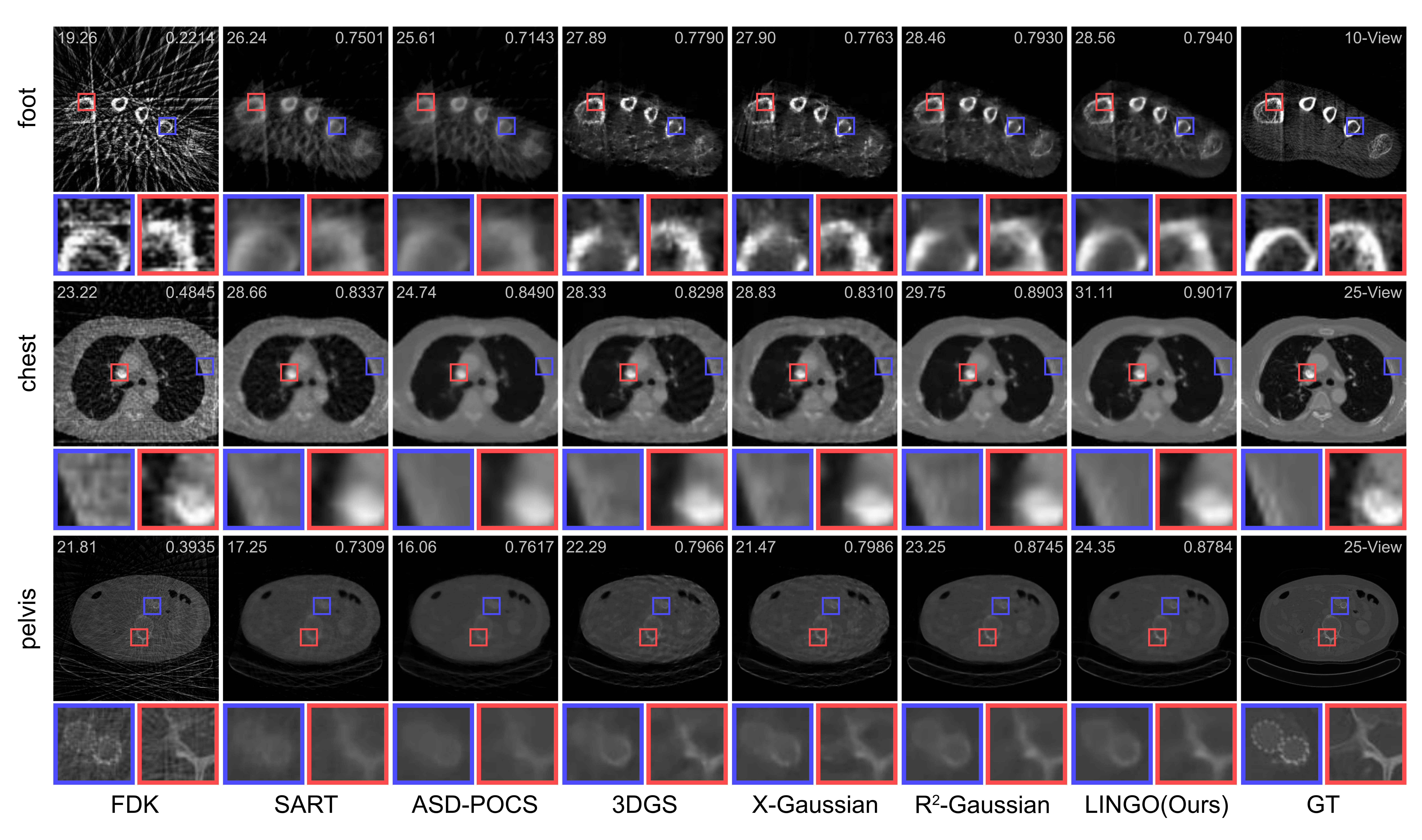}
    \caption{Visualization of results for the CT reconstruction task on X3D dataset.}
    \label{fig:visual_result_ct}
\end{figure*}

\subsection{Dynamic Gradient Optimization}

To optimize the Gaussian parameters, we utilize a loss function as follows:
\begin{equation}
\mathcal{L}_{\text{total}} 
= \mathcal{L}_1(\mathcal{I}_r, \mathcal{I}_m) 
+ \lambda_{\text{ssim}} \mathcal{L}_{\text{ssim}}(\mathcal{I}_r, \mathcal{I}_m)
+ \lambda_{\text{tv}} \mathcal{L}_{\text{tv}}(\mathbf{V}_{\text{tv}}),
\end{equation}
where $\mathcal{I}_r$ and $\mathcal{I}_m$ denote the rendered and ground-truth images, respectively. $\lambda_{\text{ssim}}$ and $\lambda_{\text{tv}}$ are the coefficients for the Structural Similarity Index (SSIM) and Total Variation (TV) regularization. $\mathbf{V}_{\text{tv}}$ represents the voxelized volumetric density field. $\mathcal{L}_{\text{tv}}(\cdot)$ minimizes its spatial gradients to encourage local smoothness and suppress noise.

To address vanishing gradients in low-density structural regions, a dynamic loss-scaling strategy is introduced. The total loss $\mathcal{L}_{\text{total}}$ is scaled by a coefficient determined using Otsu’s threshold.
\begin{equation}
	T_g = \max(0.01, \text{Otsu}(\mathcal{I} _r)).
\end{equation}
This coefficient is computed from the rendered image $I_r$ as follows:
\begin{equation}
	\mathcal{L}_{\text{final}} 
	= \mathcal{L}_{\text{total}} \times \left(T_g + \frac{1}{4T_g}\right).
\end{equation}

This scaling function is designed to strengthen gradient responses in extremely low-density areas (where \(T_g\) is small) without excessively amplifying gradients in normal-density regions, thus preventing gradient explosion. By dynamically adjusting the scaling factor based on the rendered image threshold, this strategy accelerates convergence in challenging low-density areas and improves the reconstruction of fine structures. It ultimately enhances the accuracy of the overall reconstruction.

\subsection{Initialization Point Cloud Structural Deviation}

\subsubsection{Observations and Theoretical Motivations}

During training, we observe that dynamic operations such as Gaussian splitting and cloning are primarily concentrated in structural regions. This is because these regions are consistently observed across multiple views, receiving stable and strong positive gradients that provide sufficient optimization signals. In contrast, noise points in non-structural regions suffer from gradient cancellation, leading to weak and inconsistent updates. As a result, they rarely undergo splitting or cloning and are difficult to prune effectively.

We can mathematically explain this phenomenon. For computational simplicity, let the loss function be the Mean Squared Error (MSE), and each Gaussian primitive is simplified to a Gaussian point, retaining only its density attribute. Given that the X-ray imaging model generates images based on density accumulation, we can derive the projected intensity for a pixel $k$ in view $v$ as:
\begin{equation}
  \hat{I}_{k}^v = \sum_{i=1}^{N} G_{k,i}^v \rho_i,
\end{equation}
where $G_{k,i}^v$ is the projection weight of the $i$th Gaussian point onto the $k$-th pixel of the $v$-th view satisfying $G_{k,i}^v \in \{0, 1\}$.
Let $r_{k}^v = \hat{I}_{k}^v - {I}_{k}^v$, then the MSE loss is given by:
\begin{equation}
  L_{MSE} = \frac{1}{K} \sum_{k=1}^{K}{\left(\hat{I}_k^v - {I}_k^v\right)^2 = \frac{1}{K}} \sum_{k=1}^{K} {r_{k}^v}^2, 
\end{equation}
where ${I}_k^v$ is the ground truth intensity for the $k$ pixel in view $v$, $K$ the total number of pixels in the view, $N$ the number of Gaussian points, $r_{k}^v$ the residual between the rendered and ground truth pixel. The derivative of the loss function with respect to the density $\rho_i$ of each point is:
\begin{equation}
  \frac{\partial L}{\partial \rho_i} = \frac{\partial L}{\partial r_{k}^v} \frac{\partial r_{k}^v}{\partial \rho_i} = \frac{2}{K}\sum_{k=1}^{K} r_{k}^v G_{k,i}^v.
\end{equation}
Let $G_{i}^v$ be the set of rays passing through the $i$th point in view $v$, and $|G_{i}^v|$ be the number of rays in $G_{i}^v$. Since $G_{k,i}^v=1$ for $k$th ray is in $G_{i}^v$ or else $G_{k,i}^v\neq 0$, then the gradient can be expressed as:
\begin{equation}
	\frac{\partial L}{\partial \rho_i} =\frac{2}{K}\sum_{k \in G_{i}^v} r_{k}^v.
\end{equation}
Let $S_S$s and $S_N$ denote the sets of structural points and noise points, respectively. Since more pixel rays in each view observe structural points, then $|G_{i}^v|$ will have a higher value for a structural point $s_i \in S_S$. Conversely, since very few pixel rays in each view observe noise points, it results in a much smaller $|G_{j}^v|$ for a noise point $s_j \in S_N$ that $|G_{i}^v|_{s_i \in S_N} \ll |G_{j}^v|_{s_j \in S_S}$.

From the above formulation, noise points receive gradients from only a small fraction of pixels across views, and these gradients are often unstable. As a result, they are unlikely to undergo Gaussian splitting or cloning and equally difficult to prune, since even minimal gradients from a few rays can sustain their density. In contrast, structural regions are consistently observed across multiple views and receive more stable and abundant gradients, leading to frequent Gaussian splitting and cloning in these areas.

% From the above formula, it can be observed that for noise points, even across multiple views, only a small fraction of pixel residuals contribute to the gradient, and the direction of these gradients is often unstable. Consequently, during model training, noise points are difficult to involve in Gaussian splitting and cloning, and equally difficult to prune. As long as a small number of rays provide even minimal gradients to a noise point, its density can be maintained. Conversely, structural regions exhibit consistency across multiple views and are detected by a greater number of pixel rays. This provides abundant and stable gradients for structural points during multi-view optimization, which explains why Gaussian splitting and cloning predominantly occur in structural regions.

\subsubsection{The IPSD Metric}

Building upon this theoretical understanding, we propose the Initialization Point Cloud Structural Deviation (IPSD) metric to quantify the accuracy of the initial points to the true structural distribution.

Denote the initialization point cloud by:
\begin{equation}
\mathcal{P} = \{\mathbf{p}_1, \mathbf{p}_2, \dots, \mathbf{p}_N\},
\end{equation}
and the target structural point set obtained from split/clone operations by:
\begin{equation}
\mathcal{T} = \{\mathbf{t}_1, \mathbf{t}_2, \dots, \mathbf{t}_M\}.
\end{equation}
For an arbitrary initialization point $\mathbf{p}_i$, its distance to the closest structural point as:
\begin{equation}
d(\mathbf{p}_i) = \min_{\mathbf{t} \in \mathcal{T}} \lVert \mathbf{p}_i - \mathbf{t} \rVert_2,
\end{equation}
and then we define the IPSD as:
\begin{equation}
\text{IPSD} = \frac{1}{N} \sum_{i=1}^{N} d(\mathbf{p}_i).
\end{equation}

Intuitively, IPSD measures the average distance from each initialization point to its nearest structural point discovered during training. A lower IPSD implies that the initialization point cloud is more tightly aligned with the final structures.

\section{Experiments}\label{sec:experiments}

\subsection{Experimental Setup}

\paragraph{Dataset and Hardware Environment}  
The experiments are conducted on the X3D dataset\cite{LIDC,sax_nerf}, which contains X-ray scans of various human organs, including lungs, liver, and bones. This dataset provides a solid foundation for evaluating reconstruction performance across diverse organ morphologies. To investigate the effect of training views on model performance, experiments are performed using 5, 10, and 25 X-ray images as training data. All experiments are carried out on an NVIDIA RTX 5060 Ti GPU with 16GB of VRAM.

\paragraph{Parameter Settings}  
We sample a total of 50{,}000 points with 25\% drawn from local high-density regions. To precompute the scaling ratios, we use a voxel resolution of $\mathcal{V}=(16,16,16)$. The maximum scalable voxel count $V_S$ is set to 20.  We utilize the same configuration as in $\text{R}^2$-Gaussian is adopted to maintain comparability for the loss function, where $\lambda_{\text{ssim}}$ and $\lambda_{\text{tv}}$ are set to 0.25 and 0.05. 

\paragraph{Comparison Methods and Evaluation Metrics}  
We compare our method with three explicit neural rendering models: 3DGS\cite{3dgs}, X-Gaussian\cite{x_gaussian}, and $\text{R}^2$-Gaussian\cite{r2_gaussian}. Evaluation metrics include Peak Signal-to-Noise Ratio (PSNR), SSIM, and IPSD.

\begin{figure*}[ht]
\centering
    \includegraphics[width=\textwidth]{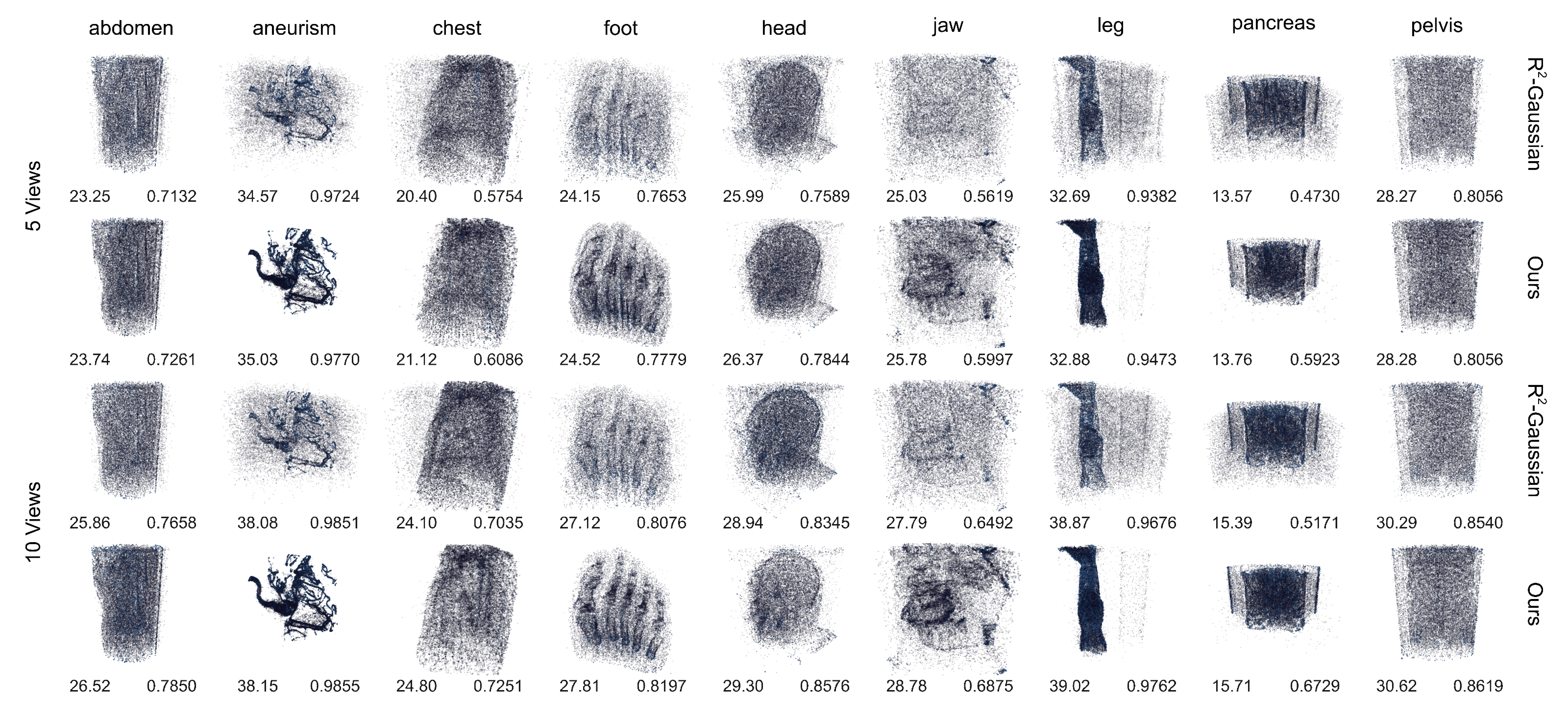}
    \caption{Point cloud structures of different models under sparse-view conditions, along with 3D PSNR and 3D SSIM metrics.}
    \label{fig:point_cloud} 
\end{figure*}

\subsection{Novel View Synthesis Task}

We evaluate four models on X-ray novel view synthesis under varying numbers of training views and iterations. Results in Table~\ref{tab:novel_view_synthesis} show that LINGO consistently achieves the best performance across all settings. Notably, under sparse-view conditions (5 and 10 views), LINGO reaches performance comparable to or exceeding the baseline trained with 30k iterations within only 5k iterations. For 25 views, LINGO within 15k iterations already surpasses the state-of-the-art model trained with 30k iterations.

The IPSD metric further shows that the initialization error is reduced from approximately 0.12 in the baseline to around 0.02 with LINGO. As illustrated in Fig.~\ref{fig:point_cloud}, FDK\cite{FDK} initialization contains substantial noise, whereas our method produces a more compact point cloud closely aligned with the ground-truth structure.

Visual comparisons in Fig.~\ref{fig:visual_result_xray} show that all methods improve as the number of views increases, while LINGO consistently delivers the best performance. Under sparse-view conditions, it better preserves structural integrity and fine details, achieving clearly superior quantitative results. In contrast, other methods exhibit noticeable degradation. They suffer from evident artifacts across all three datasets due to the inconsistency between their underlying physical models and X-ray imaging.

Although $\text{R}^2$-Gaussian alleviates this issue by incorporating X-ray physics, it remains limited by noisy initialization and insufficient denoising capability. Specifically, it produces pronounced artifacts on the structurally complex jaw dataset and shows structural distortion or missing regions on the abdomen and head datasets. In comparison, LINGO demonstrates stronger structural recovery and improved robustness against artifacts across all datasets, resulting in superior image quality and structural accuracy under sparse-view conditions.

% Visual comparisons in Fig.~\ref{fig:visual_result_xray} show that all methods improve with more views, while LINGO consistently delivers the best results. Under sparse-view conditions, it better preserves structural integrity and fine details, achieving clearly superior quantitative performance. Even with only 5 views, LINGO produces results close to ground truth.

% With 25 views, LINGO further improves to a PSNR of 47.04 and an SSIM of 0.9911, becoming nearly indistinguishable from ground truth. In contrast, other methods suffer from blurring under sparse views, and volumetric rendering methods introduce artifacts due to mismatches with X-ray physics. Although $\text{R}^2$-Gaussian improves over these methods, it remains inferior to LINGO due to noisy initialization and limited denoising capability. Overall, LINGO demonstrates superior image quality and structural accuracy under sparse-view conditions.

\begin{table*}[htbp]
\centering
\caption{Impact of different modules on quantitative metrics under 25 views.}
\begin{tabular}{|c|c|c|c|c|c|c|c|c|}
\hline
{Grad} & {Latent} & {Voxel} & {Init Time} & \multicolumn{3}{c|}{{Novel View Synthesis}} & \multicolumn{2}{c|}{{CT Reconstruction}} \\
\cline{4-9}
{Scale} & {Init} & {Scale} & {Second$\downarrow$} & {PSNR$\uparrow$} & {SSIM$\uparrow$} & {IPSD$\downarrow$} & {PSNR$\uparrow$} & {SSIM$\uparrow$} \\
\hline 
 & & & \multirow{2}{*}{1.42} & 46.09 & 0.9899 & 0.1129 & 32.61 & 0.8864 \\[0.3ex]
\cline{1-3} \cline{5-9} 
 \checkmark & & & & 46.77 & 0.9907 & 0.0906 & 32.78 & 0.8883 \\[0.3ex]
\hline
 & \checkmark & & \multirow{2}{*}{5.76} & 46.28 & 0.9899 & 0.0526 & 32.81 & 0.8911 \\[0.3ex]
\cline{1-3} \cline{5-9} 
\checkmark & \checkmark & & & 46.74 & 0.9907 & 0.0321 & 32.75 & 0.8905 \\[0.3ex]
\hline
 & \checkmark & \checkmark & \multirow{2}{*}{\textbf{0.47}} & 46.49 & 0.9902 & 0.0472 & 32.67 & 0.8894 \\[0.3ex]
\cline{1-3} \cline{5-9} 
\checkmark & \checkmark & \checkmark & & \textbf{47.04} & \textbf{0.9911} & \textbf{0.0289} & \textbf{32.91} & \textbf{0.8924} \\[0.3ex]
\hline
\end{tabular}
\label{tab:module_ablation}
\end{table*}

\subsection{CT Reconstruction Task}

We assume 100 projection views for each case in the experimental setup.  A subset of these views (e.g., 5, 10, or 25) is used as ground-truth training images, while the remaining views are synthesized via novel view synthesis and then used for CT reconstruction. For the Gaussian-based models shown in Fig.~\ref{fig:visual_result_ct}, the final CT volumes are reconstructed from the completed 100-view projections using the ASD-POCS algorithm.

Quantitative results in Table~\ref{tab:ct_reconstruction} and visual results in Fig.~\ref{fig:visual_result_ct} demonstrate that LINGO achieves consistently superior and stable performance across varying numbers of projection views. Compared with other methods, LINGO significantly improves PSNR and SSIM while maintaining strong structural fidelity even under sparse-view settings (e.g., 10 views). In particular, under identical configurations, LINGO consistently outperforms other methods across all CT reconstruction backbones. Notably, SSIM shows an improvement of approximately 0.015 over $\text{R}^2$-Gaussian, indicating that LINGO provides more accurate structural reconstruction and better preserves anatomical consistency.

From the visual results in Fig.~\ref{fig:visual_result_ct}, a similar trend can be observed across all three datasets. For volume rendering–based methods such as 3DGS and X-Gaussian, artifacts introduced in the novel view synthesis stage propagate to the CT reconstruction process, leading to more severe reconstruction errors, particularly evident in the foot dataset. Although $\text{R}^2$-Gaussian mitigates part of this issue by incorporating X-ray physics, it is still affected by noisy initialization. As a result, structural discontinuities can be observed in the foot dataset, and undesired artifacts appear in regions that should be smooth in the chest dataset. In contrast, benefiting from a more stable initialization of the Gaussian point cloud, LINGO avoids structural breakage in the foot dataset and exhibits fewer artifacts overall. In the pelvis dataset, while $\text{R}^2$-Gaussian and LINGO produce visually comparable results, LINGO achieves higher PSNR and SSIM values, indicating smaller pixel-wise discrepancies and more accurate reconstruction.

% Quantitative results in Table~\ref{tab:ct_reconstruction} and visual results in Fig.~\ref{fig:visual_result_ct} show that LINGO achieves consistently superior and stable performance across varying numbers of projection views. Compared with other methods, LINGO significantly improves PSNR and SSIM while maintaining strong structural fidelity even under sparse views (e.g., 10 views).

Compared with Gaussian-based methods, LINGO achieves the highest PSNR/SSIM, indicating superior noise suppression and edge preservation. This advantage is further supported in the point cloud domain: both point cloud quality and voxelized 3D PSNR/SSIM demonstrate that LINGO outperforms $\text{R}^2$-Gaussian in denoising and structural recovery. As shown in Fig.~\ref{fig:point_cloud}, FDK-based initialization retains substantial noise, whereas LINGO produces cleaner and more structurally consistent point clouds.

This effect is particularly evident in anatomically complex regions such as aneurism, foot, jaw, and leg. The simplified FDK initialization used in $\text{R}^2$-Gaussian preserves a large amount of noise, which is difficult to eliminate during training, ultimately leading to structural errors in both novel view synthesis and CT reconstruction. In contrast, benefiting from a more accurate and stable initialization, LINGO yields point clouds with more complete structures and significantly fewer noisy points. Notably, even under extremely sparse conditions (e.g., 5 views), LINGO achieves a level of denoising comparable to that obtained with more densely sampled views, further demonstrating its robustness and effectiveness.

% Compared with Gaussian-based methods, LINGO achieves the highest PSNR/SSIM, indicating superior noise suppression and edge preservation. This is further supported in the point cloud domain: both point cloud quality and voxelized 3D PSNR/SSIM show that LINGO outperforms $\text{R}^2$-Gaussian in denoising and structural recovery. As shown in Fig.~\ref{fig:point_cloud}, FDK initialization retains substantial noise, whereas LINGO produces cleaner and more consistent point clouds. Even with 5 views, LINGO reconstructs more complete structures with fewer artifacts.

Visual comparisons further confirm that LINGO excels in recovering tissue boundaries, low-contrast regions, and fine anatomical details, while other Gaussian-based methods tend to over-smooth or introduce artifacts. LINGO achieves a better balance between noise suppression and detail preservation. This advantage is particularly evident in local magnified regions, where structural clarity and texture fidelity are substantially improved.

\subsection{Ablation Study}

The quantitative results in Table~\ref{tab:module_ablation} show the contribution of gradient scaling, latent initialization, and voxel scaling to performance improvement. Their effects differ across tasks. Enabling any single module leads to measurable gains, but the improvements remain limited when applied in isolation.

When all three modules are enabled, the model achieves the best performance across all key metrics. In novel view synthesis, the PSNR reaches 47.04, SSIM reaches 0.9911, and IPSD attains its lowest value, indicating improved robustness to noise and reduced structural deviation. In the CT reconstruction task, the model also achieves the highest PSNR and SSIM, demonstrating consistent improvements in reconstruction quality.

Importantly, with voxel scaling enabled, the initialization speed is approximately 12 times faster than the non-scaled setting and about 3 times faster than FDK-based initialization, while maintaining comparable or improved reconstruction accuracy.

Overall, the ablation study highlights the distinct roles of the three modules: gradient scaling stabilizes optimization and enhances gradient effectiveness, latent initialization provides a more accurate structural prior, and voxel scaling reduces initialization overhead while preserving structural information. Together, these components improve both reconstruction quality and computational efficiency.

\section{Conclusion}\label{sec:conclusions}

In this work, we address the challenges of sparse-view X-ray novel view synthesis and CT reconstruction, including insufficient viewpoint information, structural blurring, and noise accumulation, by proposing the Latent Initialization and Gradient Optimization (LINGO) method. LINGO enhances the $\text{R}^2$-Gaussian framework through latent mask initialization and gradient optimization mechanisms. The key innovations include: (1) generating 3D voxel filters from X-ray image masks to remove background noise and provide accurate initialization priors; (2) dynamically scaling the loss function to strengthen gradient responses in low-density structures; and (3) introducing a voxel-space resolution scaling strategy to optimize point cloud structures. Experimental results demonstrate the superiority of LINGO in both novel view synthesis and CT reconstruction tasks, with quantitative metrics consistently outperforming existing methods, particularly under sparse-view conditions. Future directions include exploring multi-modal data fusion, adaptive voxel resolution adjustment, and real-time optimization strategies to further extend the clinical applicability of LINGO.

\bibliographystyle{plain}
\bibliography{reference}

@article{SART,
	title = {Simultaneous Algebraic Reconstruction Technique (SART): A superior implementation of the ART algorithm},
	journal = {Ultrasonic Imaging},
	volume = {6},
	number = {1},
	pages = {81-94},
	year = {1984},
	issn = {0161-7346},
	doi = {https://doi.org/10.1016/0161-7346(84)90008-7},
	url = {https://www.sciencedirect.com/science/article/pii/0161734684900087},
	author = {A.H. Andersen and A.C. Kak},
}

@article{LIDC,
  author    = {Armato, Samuel G. and McLennan, Geoffrey and Bidaut, Luc and McNitt-Gray, Michael F. and Meyer, Charles R. and Reeves, Anthony P. and Zhao, Binsheng and Aberle, Denise R, et al},
  title     = {The Lung Image Database Consortium (LIDC) and Image Database Resource Initiative (IDRI): A Completed Reference Database of Lung Nodules on CT Scans},
  journal   = {Medical Physics},
  volume    = {38},
  number    = {2},
  pages     = {915--931},
  year      = {2011},
  doi       = {10.1118/1.3528204},
  pmid      = {21452728},
  pmcid     = {PMC3041807}
}

@inproceedings{NeRF,
 title={NeRF: Representing Scenes as Neural Radiance Fields for View Synthesis},
 author={Ben Mildenhall and Pratul P. Srinivasan and Matthew Tancik and Jonathan T. Barron and Ravi Ramamoorthi and Ren Ng},
 year={2020},
 booktitle={ECCV},
}

@inproceedings{TensoRF,
author = {Chen, Anpei and Xu, Zexiang and Geiger, Andreas and Yu, Jingyi and Su, Hao},
title = {TensoRF: Tensorial Radiance Fields},
year = {2022},
isbn = {978-3-031-19823-6},
publisher = {Springer-Verlag},
address = {Berlin, Heidelberg},
url = {https://doi.org/10.1007/978-3-031-19824-3_20},
doi = {10.1007/978-3-031-19824-3_20},
booktitle = {Computer Vision – ECCV 2022: 17th European Conference, Tel Aviv, Israel, October 23–27, 2022, Proceedings, Part XXXII},
pages = {333–350},
numpages = {18},
location = {Tel Aviv, Israel}
}

@article{Eff-NeRF,
  title={EfficientNeRF: Efficient Neural Radiance Fields},
  author={T. Hu and Shu Liu and Yilun Chen and Tiancheng Shen and Jiaya Jia},
  journal={ArXiv},
  year={2022},
  volume={abs/2206.00878},
  url={https://api.semanticscholar.org/CorpusID:249282202}
}

@INPROCEEDINGS{Mip-NeRF,
  author={Barron, Jonathan T. and Mildenhall, Ben and Tancik, Matthew and Hedman, Peter and Martin-Brualla, Ricardo and Srinivasan, Pratul P.},
  booktitle={2021 IEEE/CVF International Conference on Computer Vision (ICCV)}, 
  title={Mip-NeRF: A Multiscale Representation for Anti-Aliasing Neural Radiance Fields}, 
  year={2021},
  volume={},
  number={},
  pages={5835-5844},
  doi={10.1109/ICCV48922.2021.00580}}

@article{Mip-NeRF360,
  title={Mip-NeRF 360: Unbounded Anti-Aliased Neural Radiance Fields},
  author={Jonathan T. Barron and Ben Mildenhall and Dor Verbin and Pratul P. Srinivasan and Peter Hedman},
  journal={2022 IEEE/CVF Conference on Computer Vision and Pattern Recognition (CVPR)},
  year={2021},
  pages={5460-5469},
  url={https://api.semanticscholar.org/CorpusID:244488448}
}

@article{Zip-NeRF,
    title={Zip-NeRF: Anti-Aliased Grid-Based Neural Radiance Fields},
    author={Jonathan T. Barron and Ben Mildenhall and 
            Dor Verbin and Pratul P. Srinivasan and Peter Hedman},
    journal={ICCV},
    year={2023}
}

@article{Med-NeRF,
  title={MedNeRF: Medical Neural Radiance Fields for Reconstructing 3D-aware CT-Projections from a Single X-ray},
  author={Abril Corona-Figueroa and Jonathan Frawley and Sam Bond-Taylor and Sarath Bethapudi and Hubert P. H. Shum and Chris G. Willcocks},
  journal={2022 44th Annual International Conference of the IEEE Engineering in Medicine \& Biology Society (EMBC)},
  year={2022},
  pages={3843-3848},
  url={https://api.semanticscholar.org/CorpusID:246473192}
}

@inproceedings{NAF,
	author = {Zha, Ruyi and Zhang, Yanhao and Li, Hongdong},
	title = {NAF: Neural Attenuation Fields for nbsp;Sparse-View CBCT Reconstruction},
	year = {2022},
	isbn = {978-3-031-16445-3},
	publisher = {Springer-Verlag},
	address = {Berlin, Heidelberg},
	url = {https://doi.org/10.1007/978-3-031-16446-0_42},
	doi = {10.1007/978-3-031-16446-0_42},
	booktitle = {Medical Image Computing and Computer Assisted Intervention – MICCAI 2022: 25th International Conference, Singapore, September 18–22, 2022, Proceedings, Part VI},
	pages = {442–452},
	numpages = {11},
	location = {Singapore, Singapore}
}

@inproceedings{sax_nerf,
  title={Structure-Aware Sparse-View X-ray 3D Reconstruction},
  author={Yuanhao Cai and Jiahao Wang and Alan Yuille and Zongwei Zhou and Angtian Wang},
  booktitle={CVPR},
  year={2024}
}

@Article{3dgs,
      author       = {Kerbl, Bernhard and Kopanas, Georgios and Leimk{\"u}hler, Thomas and Drettakis, George},
      title        = {3D Gaussian Splatting for Real-Time Radiance Field Rendering},
      journal      = {ACM Transactions on Graphics},
      number       = {4},
      volume       = {42},
      month        = {July},
      year         = {2023},
      url          = {https://repo-sam.inria.fr/fungraph/3d-gaussian-splatting/}
}

@inproceedings{x_gaussian,
  title={Radiative gaussian splatting for efficient x-ray novel view synthesis},
  author={Yuanhao Cai and Yixun Liang and Jiahao Wang and Angtian Wang and Yulun Zhang and Xiaokang Yang and Zongwei Zhou and Alan Yuille},
  booktitle={ECCV},
  year={2024}
}

@inproceedings{r2_gaussian,
  title={R2-Gaussian: Rectifying Radiative Gaussian Splatting for Tomographic Reconstruction},
  author={Ruyi Zha and Tao Jun Lin and Yuanhao Cai and Jiwen Cao and Yanhao Zhang and Hongdong Li},
  booktitle={NeurIPS},
  year={2024}
}

@article{LucidDreamer,
  title={LucidDreamer: Towards High-Fidelity Text-to-3D Generation via Interval Score Matching},
  author={Yixun Liang and Xin Yang and Jiantao Lin and Haodong Li and Xiaogang Xu and Yingcong Chen},
  journal={2024 IEEE/CVF Conference on Computer Vision and Pattern Recognition (CVPR)},
  year={2023},
  pages={6517-6526},
  url={https://api.semanticscholar.org/CorpusID:265295106}
}

@ARTICLE{ct-tech,
	author = {{Cormack}, A.~M.},
	title = "{Representation of a Function by Its Line Integrals, with Some Radiological Applications}",
	journal = {Journal of Applied Physics},
	year = 1963,
	month = sep,
	volume = {34},
	number = {9},
	pages = {2722-2727},
	doi = {10.1063/1.1729798},
	adsurl = {https://ui.adsabs.harvard.edu/abs/1963JAP....34.2722C}
}

@article{HUGSHG,
  title={HUGS: Human Gaussian Splats},
  author={Muhammed Kocabas and Jen-Hao Rick Chang and James Gregory Gabriel and Oncel Tuzel and Anurag Ranjan},
  journal={2024 IEEE/CVF Conference on Computer Vision and Pattern Recognition (CVPR)},
  year={2023},
  pages={505-515},
  url={https://api.semanticscholar.org/CorpusID:265498305}
}

@inproceedings {EWA,
    title={EWA volume splatting},
    author={Zwicker, Matthias and Pfister, Hanspeter and Van Baar, Jeroen and Gross, Markus},
    booktitle={Visualization, 2001. VIS 01. Proceedings},
    pages={29--538},
    year={2001},
    organization={IEEE}
}

@misc{humanarticulatedgaussiansplatting,
      title={GauHuman: Articulated Gaussian Splatting from Monocular Human Videos}, 
      author={Shoukang Hu and Ziwei Liu},
      year={2023},
      eprint={2312.02973},
      archivePrefix={arXiv},
      primaryClass={cs.CV},
      url={https://arxiv.org/abs/2312.02973}, 
}

@article{FDK,
  author={Hsieh, J.},
  booktitle={2000 IEEE Nuclear Science Symposium. Conference Record (Cat. No.00CH37149)}, 
  title={A practical cone beam artifact correction algorithm}, 
  year={2000},
  volume={2},
  number={},
  pages={15/71-15/74 vol.2},
  doi={10.1109/NSSMIC.2000.950053}}

@inproceedings{DDRs,
 author = {Gao, Zhongpai and Planche, Benjamin and Zheng, Meng and Chen, Xiao and Chen, Terrence and Wu, Ziyan},
 booktitle = {Advances in Neural Information Processing Systems},
 doi = {10.52202/079017-1240},
 editor = {A. Globerson and L. Mackey and D. Belgrave and A. Fan and U. Paquet and J. Tomczak and C. Zhang},
 pages = {39281--39302},
 publisher = {Curran Associates, Inc.},
 title = {DDGS-CT: Direction-Disentangled Gaussian Splatting for Realistic Volume Rendering},
 url = {https://proceedings.neurips.cc/paper_files/paper/2024/file/456ce0476c9b4689a74918b851cecd5a-Paper-Conference.pdf},
 volume = {37},
 year = {2024}
}

@article{low-doseCT,
  title={Deep Learning-Based Sinogram Completion for Low-Dose CT},
  author={Muhammad Usman Ghani and W. Clem Karl},
  journal={2018 IEEE 13th Image, Video, and Multidimensional Signal Processing Workshop (IVMSP)},
  year={2018},
  pages={1-5},
  url={https://api.semanticscholar.org/CorpusID:52148333}
}

@article{gs2mesh,
  title={SuGaR: Surface-Aligned Gaussian Splatting for Efficient 3D Mesh Reconstruction and High-Quality Mesh Rendering},
  author={Antoine Gu{\'e}don and Vincent Lepetit},
  journal={2024 IEEE/CVF Conference on Computer Vision and Pattern Recognition (CVPR)},
  year={2023},
  pages={5354-5363},
  url={https://api.semanticscholar.org/CorpusID:265308825}
}

@article{ct-med,
	author = {Godfrey N. Hounsfield },
	title = {Computed Medical Imaging},
	journal = {Science},
	volume = {210},
	number = {4465},
	pages = {22-28},
	year = {1980},
	doi = {10.1126/science.6997993},
	URL = {https://www.science.org/doi/abs/10.1126/science.6997993},
	eprint = {https://www.science.org/doi/pdf/10.1126/science.6997993}
}

@article{Dynamic3G,
  title={Dynamic 3D Gaussians: Tracking by Persistent Dynamic View Synthesis},
  author={Jonathon Luiten and Georgios Kopanas and Bastian Leibe and Deva Ramanan},
  journal={2024 International Conference on 3D Vision (3DV)},
  year={2023},
  pages={800-809},
  url={https://api.semanticscholar.org/CorpusID:261030923}
}

@book{ct-principle,
	author = {Kak, Avinash C. and Slaney, Malcolm},
	title = {Principles of Computerized Tomographic Imaging},
	publisher = {Society for Industrial and Applied Mathematics},
	year = {2001},
	doi = {10.1137/1.9780898719277},
	address = {},
	edition   = {},
	URL = {https://epubs.siam.org/doi/abs/10.1137/1.9780898719277},
	eprint = {https://epubs.siam.org/doi/pdf/10.1137/1.9780898719277}
}

@article{SparseView2DCT,
	title = {3DGR-CT: Sparse-view CT reconstruction with a 3D Gaussian representation},
	journal = {Medical Image Analysis},
	volume = {103},
	pages = {103585},
	year = {2025},
	issn = {1361-8415},
	doi = {https://doi.org/10.1016/j.media.2025.103585},
	url = {https://www.sciencedirect.com/science/article/pii/S136184152500132X},
	author = {Yingtai Li and Xueming Fu and Han Li and Shang Zhao and Ruiyang Jin and S. Kevin Zhou}
}

@article{low-dose-GAN,
  author={Wolterink, Jelmer M. and Leiner, Tim and Viergever, Max A. and Išgum, Ivana},
  journal={IEEE Transactions on Medical Imaging}, 
  title={Generative Adversarial Networks for Noise Reduction in Low-Dose CT}, 
  year={2017},
  volume={36},
  number={12},
  pages={2536-2545},
  doi={10.1109/TMI.2017.2708987}
}

@article{TomoGAN,
  author       = {Liu, Zhengchun and Bicer, Tekin and Kettimuthu, Rajkumar and Gursoy, Doga and De Carlo, Francesco and Foster, Ian},
  title        = {TomoGAN: low-dose synchrotron x-ray tomography with generative adversarial networks: discussion},
  doi          = {10.1364/JOSAA.375595},
  url          = {https://www.osti.gov/biblio/1635605},
  journal      = {Journal of the Optical Society of America. A, Optics, Image Science, and Vision},
  issn         = {ISSN 1084-7529},
  number       = {3},
  volume       = {37},
  place        = {United States},
  publisher    = {Optical Society of America (OSA)},
  year         = {2020},
  month        = {02}
}

@article{GSIR,
  title={GS-IR: 3D Gaussian Splatting for Inverse Rendering},
  author={Zhihao Liang and Qi Zhang and Yingfa Feng and Ying Shan and Kui Jia},
  journal={2024 IEEE/CVF Conference on Computer Vision and Pattern Recognition (CVPR)},
  year={2023},
  pages={21644-21653},
  url={https://api.semanticscholar.org/CorpusID:265466404}
}

@INPROCEEDINGS{NerfAcc,
  author={Li, Ruilong and Gao, Hang and Tancik, Matthew and Kanazawa, Angjoo},
  booktitle={2023 IEEE/CVF International Conference on Computer Vision (ICCV)}, 
  title={NerfAcc: Efficient Sampling Accelerates NeRFs}, 
  year={2023},
  volume={},
  number={},
  pages={18491-18500},
  doi={10.1109/ICCV51070.2023.01699}
}

@ARTICLE{Otsu,
  author={Otsu, Nobuyuki},
  journal={IEEE Transactions on Systems, Man, and Cybernetics}, 
  title={A Threshold Selection Method from Gray-Level Histograms}, 
  year={1979},
  volume={9},
  number={1},
  pages={62-66},
  doi={10.1109/TSMC.1979.4310076}}

@misc{slam,
      title={Gaussian Splatting SLAM}, 
      author={Hidenobu Matsuki and Riku Murai and Paul H. J. Kelly and Andrew J. Davison},
      year={2024},
      eprint={2312.06741},
      archivePrefix={arXiv},
      primaryClass={cs.CV},
      url={https://arxiv.org/abs/2312.06741}, 
}

@INPROCEEDINGS{colmap,
  author={Schönberger, Johannes L. and Frahm, Jan-Michael},
  booktitle={2016 IEEE Conference on Computer Vision and Pattern Recognition (CVPR)}, 
  title={Structure-from-Motion Revisited}, 
  year={2016},
  volume={},
  number={},
  pages={4104-4113},
  doi={10.1109/CVPR.2016.445}
}

@article{MERF,
author = {Reiser, Christian and Szeliski, Rick and Verbin, Dor and Srinivasan, Pratul and Mildenhall, Ben and Geiger, Andreas and Barron, Jon and Hedman, Peter},
title = {MERF: Memory-Efficient Radiance Fields for Real-time View Synthesis in Unbounded Scenes},
year = {2023},
issue_date = {August 2023},
publisher = {Association for Computing Machinery},
address = {New York, NY, USA},
volume = {42},
number = {4},
issn = {0730-0301},
url = {https://doi.org/10.1145/3592426},
doi = {10.1145/3592426},
journal = {ACM Trans. Graph.},
month = jul,
articleno = {89},
numpages = {12}
}

@ARTICLE{os,
  author={Hudson, H.M. and Larkin, R.S.},
  journal={IEEE Transactions on Medical Imaging}, 
  title={Accelerated image reconstruction using ordered subsets of projection data}, 
  year={1994},
  volume={13},
  number={4},
  pages={601-609},
  doi={10.1109/42.363108}
}

@article{ASD-POCS,
	doi = {10.1088/0031-9155/53/17/021},
	url = {https://doi.org/10.1088/0031-9155/53/17/021},
	year = {2008},
	month = {aug},
	publisher = {},
	volume = {53},
	number = {17},
	pages = {4777},
	author = {Sidky, Emil Y and Pan, Xiaochuan},
	title = {Image reconstruction in circular cone-beam computed tomography by constrained, total-variation minimization},
	journal = {Physics in Medicine Biology}
}

@inproceedings{GaussianShader,
	title = "GaussianShader: 3D Gaussian Splatting with Shading Functions for Reflective Surfaces",
	author = "Yingwenqi Jiang and Jiadong Tu and Yuan Liu and Xifeng Gao and Xiaoxiao Long and Wenping Wang and Yuexin Ma",
	year = "2024",
	doi = "10.1109/CVPR52733.2024.00509",
	language = "English",
	series = "Proceedings of the IEEE Computer Society Conference on Computer Vision and Pattern Recognition",
	publisher = "IEEE Computer Society",
	pages = "5322--5332",
	booktitle = "Proceedings - 2024 IEEE/CVF Conference on Computer Vision and Pattern Recognition, CVPR 2024",
}

@article{cgls,
	author = {Paige, Christopher C. and Saunders, Michael A.},
	title = {LSQR: An Algorithm for Sparse Linear Equations and Sparse Least Squares},
	year = {1982},
	issue_date = {March 1982},
	publisher = {Association for Computing Machinery},
	address = {New York, NY, USA},
	volume = {8},
	number = {1},
	issn = {0098-3500},
	url = {https://doi.org/10.1145/355984.355989},
	doi = {10.1145/355984.355989},
	journal = {ACM Trans. Math. Softw.},
	month = mar,
	pages = {43–71},
	numpages = {29}
}

@ARTICLE{ssim,
  author={Zhou Wang and Bovik, A.C. and Sheikh, H.R. and Simoncelli, E.P.},
  journal={IEEE Transactions on Image Processing}, 
  title={Image quality assessment: from error visibility to structural similarity}, 
  year={2004},
  volume={13},
  number={4},
  pages={600-612},
  doi={10.1109/TIP.2003.819861}
}

@InProceedings{4dgs,
    author    = {Wu, Guanjun and Yi, Taoran and Fang, Jiemin and Xie, Lingxi and Zhang, Xiaopeng and Wei, Wei and Liu, Wenyu and Tian, Qi and Wang, Xinggang},
    title     = {4D Gaussian Splatting for Real-Time Dynamic Scene Rendering},
    booktitle = {Proceedings of the IEEE/CVF Conference on Computer Vision and Pattern Recognition (CVPR)},
    month     = {June},
    year      = {2024},
    pages     = {20310-20320}
}

@article{PhysGaussian,
  title={PhysGaussian: Physics-Integrated 3D Gaussians for Generative Dynamics},
  author={Tianyi Xie and Zeshun Zong and Yuxing Qiu and Xuan Li and Yutao Feng and Yin Yang and Chenfanfu Jiang},
  journal={2024 IEEE/CVF Conference on Computer Vision and Pattern Recognition (CVPR)},
  year={2023},
  pages={4389-4398},
  url={https://api.semanticscholar.org/CorpusID:265309217}
}

@article{4dgs-Photorealistic,
  title={Real-time Photorealistic Dynamic Scene Representation and Rendering with 4D Gaussian Splatting},
  author={Zeyu Yang and Hongye Yang and Zijie Pan and Xiatian Zhu and Li Zhang},
  journal={ArXiv},
  year={2023},
  volume={abs/2310.10642},
  url={https://api.semanticscholar.org/CorpusID:264172721}
}

@conference{GS-SLAM,
	title = "GS-SLAM: Dense Visual SLAM with 3D Gaussian Splatting",
	author = "Delin Qu and Chi Yan and Bin Zhao and Xuelong Li and Zhigang Wang and Dong Wang and Dan Xu",
	year = "2024",
	month = jun,
	doi = "10.1109/CVPR52733.2024.01853",
	language = "English",
	pages = "19595--19604",
	note = "Proceedings of the IEEE/CVF Conference on Computer Vision and Pattern Recognition (CVPR) ; Conference date: 01-06-2024 Through 01-06-2024",
}

@article{ct-reconstruction,
	title = "Region of interest reconstruction from truncated data in circular cone-beam CT",
	author = "Lifeng Yu and Yu Zou and Sidky, \{Emil Y.\} and Pelizzari, \{Charles A.\} and Peter Munro and Xiaochuan Pan",
	year = "2006",
	month = jul,
	doi = "10.1109/TMI.2006.872329",
	language = "English (US)",
	volume = "25",
	pages = "869--881",
	journal = "IEEE transactions on medical imaging",
	issn = "0278-0062",
	publisher = "Institute of Electrical and Electronics Engineers Inc.",
	number = "7",
}

@article{Scaffold-GS,
  title={Scaffold-GS: Structured 3D Gaussians for View-Adaptive Rendering},
  author={Tao Lu and Mulin Yu and Linning Xu and Yuanbo Xiangli and Limin Wang and Dahua Lin and Bo Dai},
  journal={2024 IEEE/CVF Conference on Computer Vision and Pattern Recognition (CVPR)},
  year={2023},
  pages={20654-20664},
  url={https://api.semanticscholar.org/CorpusID:265551778}
}

@article{alpha-blending,
	author = {Porter, Thomas and Duff, Tom},
	title = {Compositing digital images},
	year = {1984},
	issue_date = {July 1984},
	publisher = {Association for Computing Machinery},
	address = {New York, NY, USA},
	volume = {18},
	number = {3},
	issn = {0097-8930},
	url = {https://doi.org/10.1145/964965.808606},
	doi = {10.1145/964965.808606},
	journal = {SIGGRAPH Comput. Graph.},
	month = jan,
	pages = {253–259},
	numpages = {7}
}

@article{cone-beam-ct,
  title={Clinical applications of cone-beam computed tomography in dental practice.},
  author={William C. Scarfe and Allan G. Farman and Predag Sukovic},
  journal={Journal},
  year={2006},
  volume={72 1},
  pages={
          75-80
        },
  url={https://api.semanticscholar.org/CorpusID:246533}
}

\end{document}